\documentclass[conference]{IEEEtran}
\IEEEoverridecommandlockouts
\usepackage{cite}
\usepackage{amsmath,amssymb,amsfonts}
\usepackage{graphicx}
\usepackage{textcomp}
\usepackage{xcolor}
\usepackage{xurl}
\usepackage{booktabs}
\usepackage{algpseudocode}
\usepackage{enumerate}
\usepackage[english]{babel}
\usepackage{blindtext}
\usepackage{xspace}
\usepackage{multirow}
\usepackage{threeparttable}
\usepackage{float}
\usepackage{flafter}
\usepackage{comment}
\usepackage[skip=12pt]{caption}
\usepackage{graphicx}
\usepackage{adjustbox}
\usepackage{caption}
\usepackage[justification=centering]{caption}
\usepackage[ruled,linesnumbered]{algorithm2e}
\usepackage{tikz}
\usepackage{fp}
\usepackage{algpseudocode}
\usepackage{ragged2e}
\usepackage{mathrsfs}
\usepackage{amsthm}
\usepackage{colortbl}
\usepackage{wasysym}
\usepackage{fontawesome}
\usepackage{subfig}

\newlength\maxlentime
\newcommand\timebar[3][red!20]{%
  \FPeval\result{round((#3-1.5)/#2:4)}%
  \rlap{\textcolor{#1}{\hspace*{\dimexpr-\tabcolsep+.5\arrayrulewidth}%
        \rule[-.05\ht\strutbox]{\result\maxlentime}{.95\ht\strutbox}}}%
  \makebox[\dimexpr\maxlentime-0.2\tabcolsep+\arrayrulewidth][r]{#3}}

\def\headertime{New}
\def\fig{Fig.\xspace}

\def\tab{Tab.\xspace}

\newcommand{\head}[1]{{\noindent \textbf{#1:}}}

\graphicspath{{./figures/}}
\graphicspath{{./pdf/}}
\DeclareGraphicsExtensions{.pdf,.jpeg,.png}

\ifodd 1
\newcommand{\com}[1]{\textbf{\color{red}(COMMENT: #1)}} 
\newcommand{\todo}[1]{\textbf{{\color{orange}(TODO: #1)}}}
\else

\newcommand{\com}[1]{}
\newcommand{\todo}[1]{}
\fi

\def\BibTeX{{\rm B\kern-.05em{\sc i\kern-.025em b}\kern-.08em
    T\kern-.1667em\lower.7ex\hbox{E}\kern-.125emX}}
\begin{document}
\DeclareRobustCommand*{\IEEEauthorrefmark}[1]{%
  \raisebox{0pt}[0pt][0pt]{\textsuperscript{\footnotesize #1}}%
}
\title{Are You Being Tracked? Discover the Power of Zero-Shot Trajectory Tracing with LLMs!
}
\author{\IEEEauthorblockN{Huanqi Yang\IEEEauthorrefmark{1},
Sijie Ji\IEEEauthorrefmark{2},
Rucheng Wu\IEEEauthorrefmark{1},
Weitao Xu\IEEEauthorrefmark{1}
}
\IEEEauthorblockA{
\IEEEauthorrefmark{1}City University of Hong Kong
}
\IEEEauthorblockA{
\IEEEauthorrefmark{2}University of Hong Kong
}
}

\maketitle

\begin{abstract}
There is a burgeoning discussion around the capabilities of Large Language Models (LLMs) in acting as fundamental components that can be seamlessly incorporated into Artificial Intelligence of Things (AIoT) to interpret complex trajectories. This study introduces LLMTrack, a model that illustrates how LLMs can be leveraged for Zero-Shot Trajectory Recognition by employing a novel single-prompt technique that combines role-play and think step-by-step methodologies with unprocessed Inertial Measurement Unit (IMU) data. We evaluate the model using real-world datasets designed to challenge it with distinct trajectories characterized by indoor and outdoor scenarios. In both test scenarios, LLMTrack not only meets but exceeds the performance benchmarks set by traditional machine learning approaches and even contemporary state-of-the-art deep learning models, all without the requirement of training on specialized datasets. The results of our research suggest that, with strategically designed prompts, LLMs can tap into their extensive knowledge base and are well-equipped to analyze raw sensor data with remarkable effectiveness.
\end{abstract}

\begin{IEEEkeywords}
Large Language Models, AIoT, Tracking, Trajectory
\end{IEEEkeywords}

\section{Introduction}
Recent advancements in Large Language Models (LLMs) have showcased remarkable capabilities~\cite{li2024personal,zhang2024large,li2024survey}, particularly evident in instances such as Sora~\cite{videoworldsimulators2024} and Claude-3~\cite{claude2023}. Sora exemplifies how LLMs can intuitively grasp a wide range of world phenomena, from the principles of physics to the complexities of societal and artistic concepts, simulating a child's learning process. These LLMs not only excel in a diverse array of downstream tasks but also demonstrate high-level human-like reasoning and exceptional generalization abilities.
The potential of LLMs extends to the realm of the Artificial Intelligence of Things (AIoT), where such models can be utilized to enhance the intelligence of interconnected devices, enabling them to interact with and understand the world in more sophisticated and refined ways.

\begin{figure}[t]
  \begin{center}
  \includegraphics[width=0.49\textwidth]{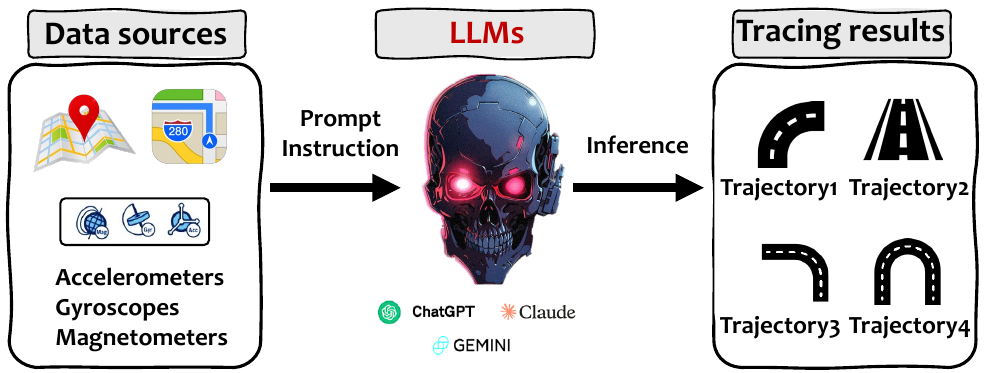}
  \caption{Workflow of LLMTrack.}\label{fig:workflow}
  \end{center}
  \vspace{-0.9cm}
\end{figure}

Despite these notable achievements, certain scholars argue that existing LLMs have merely a superficial grasp of knowledge and remain a considerable distance from attaining a profound comprehension of the physical realm~\cite{lecun2022path}. This shortcoming is often attributed to their reliance on training using extensive collections of textual and visual content from the internet, which may result in less-than-optimal performance when evaluating digital and temporal data sequences.
Moreover, these models often face challenges in meaningfully interacting with the real world, such as when they are assigned to generate precise and efficient control strategies.

The lively discussion around this topic has fueled a growing interest in examining how LLMs could become core components that are intricately interwoven with Cyber-Physical Systems (CPS) to decode aspects of the physical environment. Specifically, some researchers have introduced the notion of HARGPT~\cite{ji2024hargpt}, which investigates the potential of employing LLMs for human activity recognition, yielding promising results. These findings suggest a valuable application of LLMs in understanding and interpreting human behaviors through data analysis, indicating a significant advancement in the field. However, the full scope of what LLMs might achieve in terms of modeling and understanding the physical world in other tasks has not been thoroughly investigated.

Inspired by the strides made with LLMs and the prospects of HARGPT, this paper sets out by focusing on the fundamental application of trajectory tracing within Cyber-Physical Systems. It seeks to investigate and understand the existing abilities of LLMs by inputting unprocessed sensor data coupled with a basic prompt.
As Shown in Fig.~\ref{fig:workflow}, the unprocessed IoT sensor data is input into several renowned LLMs including ChatGPT, Claude, and Gemini, resulting in the models producing trajectory tracking results.

\begin{figure*}[t]
  \begin{center}
  \includegraphics[width=0.89\textwidth]{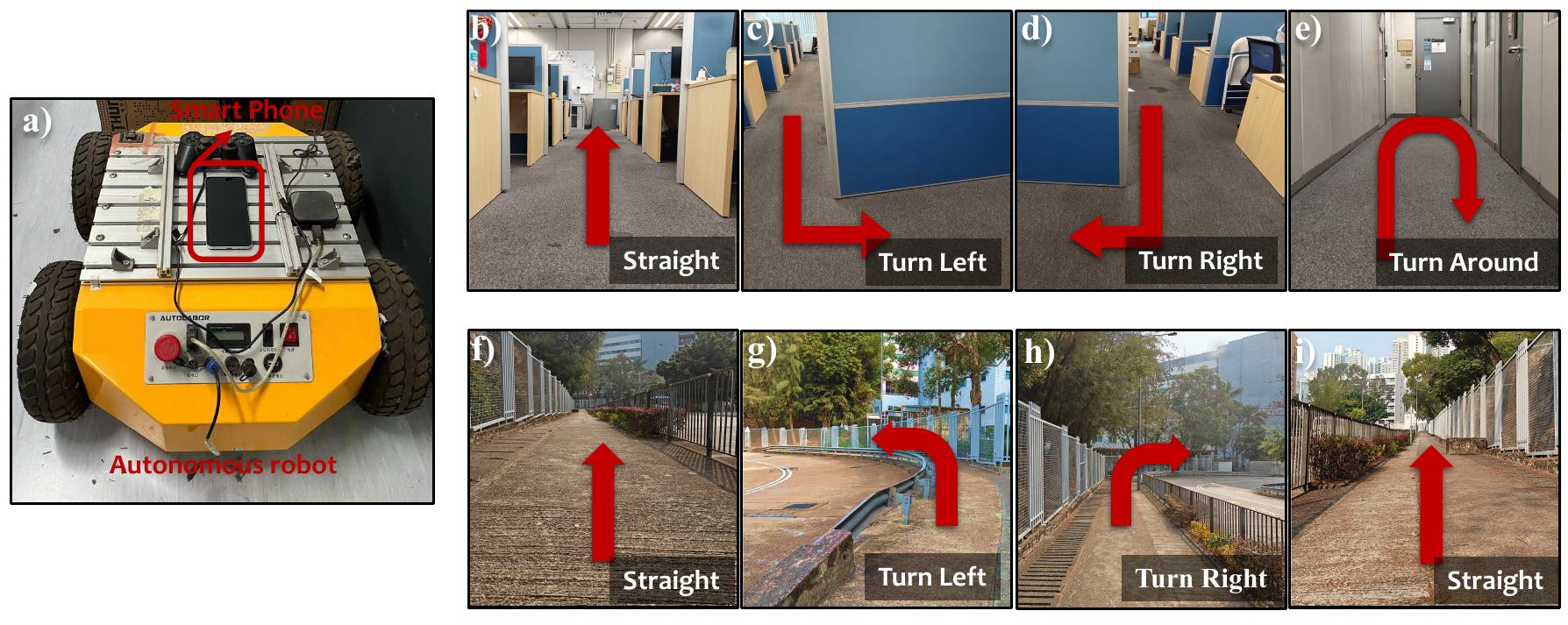}
  \caption{Experimental settings. a) Experiment devices. b)--i) Eight different scenes in indoor and outdoor environments.}\label{fig:exp}
  \end{center}
  \vspace{-0.5cm}
\end{figure*}

The performance evaluation spans two different environments: indoor and outdoor, with each having four unique types of trajectories. The benchmarks set for comparison include both traditional machine learning models and contemporary state-of-the-art deep learning models. The experimental findings reveal that LLMs possess the capability to execute zero-shot tracking with raw sensor data effectively. When compared to traditional methods, not only do LLMs surpass them, but they also attain an average accuracy above 80\%. It is particularly noteworthy that, in contrast to learning-based models which tend to suffer from performance drops when dealing with novel data and usually require retraining or fine-tuning for particular datasets, LLMs demonstrate considerable resilience.
We introduce LLMTrack and highlight the following points:
\begin{itemize}
    \item 
    LLMTrack uncovers the ability of LLMs to act as zero-shot Trajectory Tracers without the necessity for any fine-tuning or the use of prompts crafted with domain-specific knowledge.
    \item LLMTrack affirms the adeptness of LLMs in interpreting IoT sensor data and executing tasks within the physical realm.
\end{itemize}
Moreover, we engage in an in-depth analysis and discussion about the insights and significant findings from employing LLMs with IoT sensor data, which sets the stage for further studies on the integration of LLMs into AIoT systems.


\section{LLMTrack: zero-shot tracking with LLMs}
\label{sec:design}
\subsection{Experiments}

\begin{figure*}[h]
\centering
   \subfloat[Indoor turn around accelerations]{
    \includegraphics[width=0.3\textwidth]{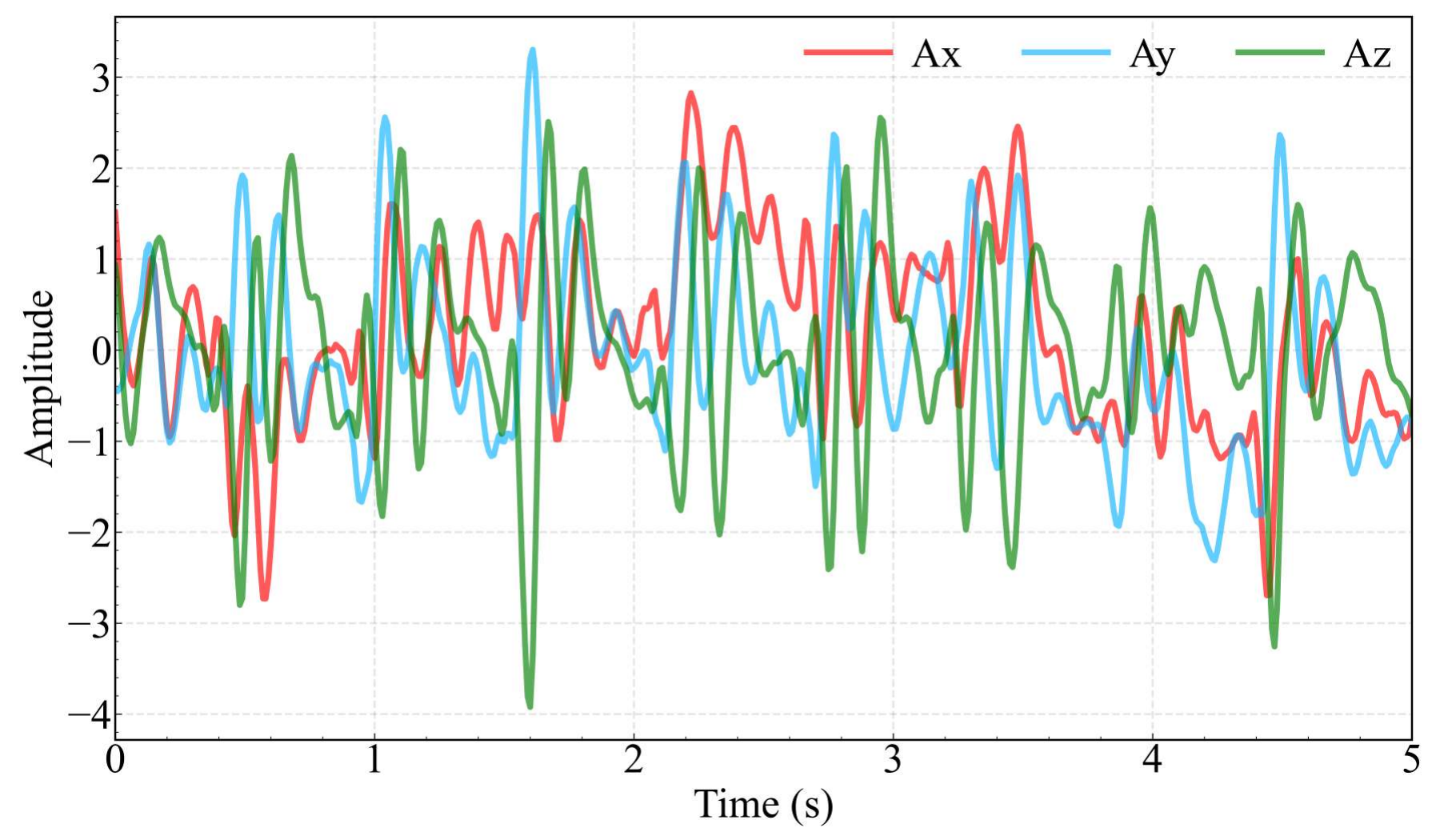}
        \label{fig:indoor1}    }
   \subfloat[Indoor turn around gyroscopes]{
    \label{fig:indoor2}
    \includegraphics[width=0.3\textwidth]{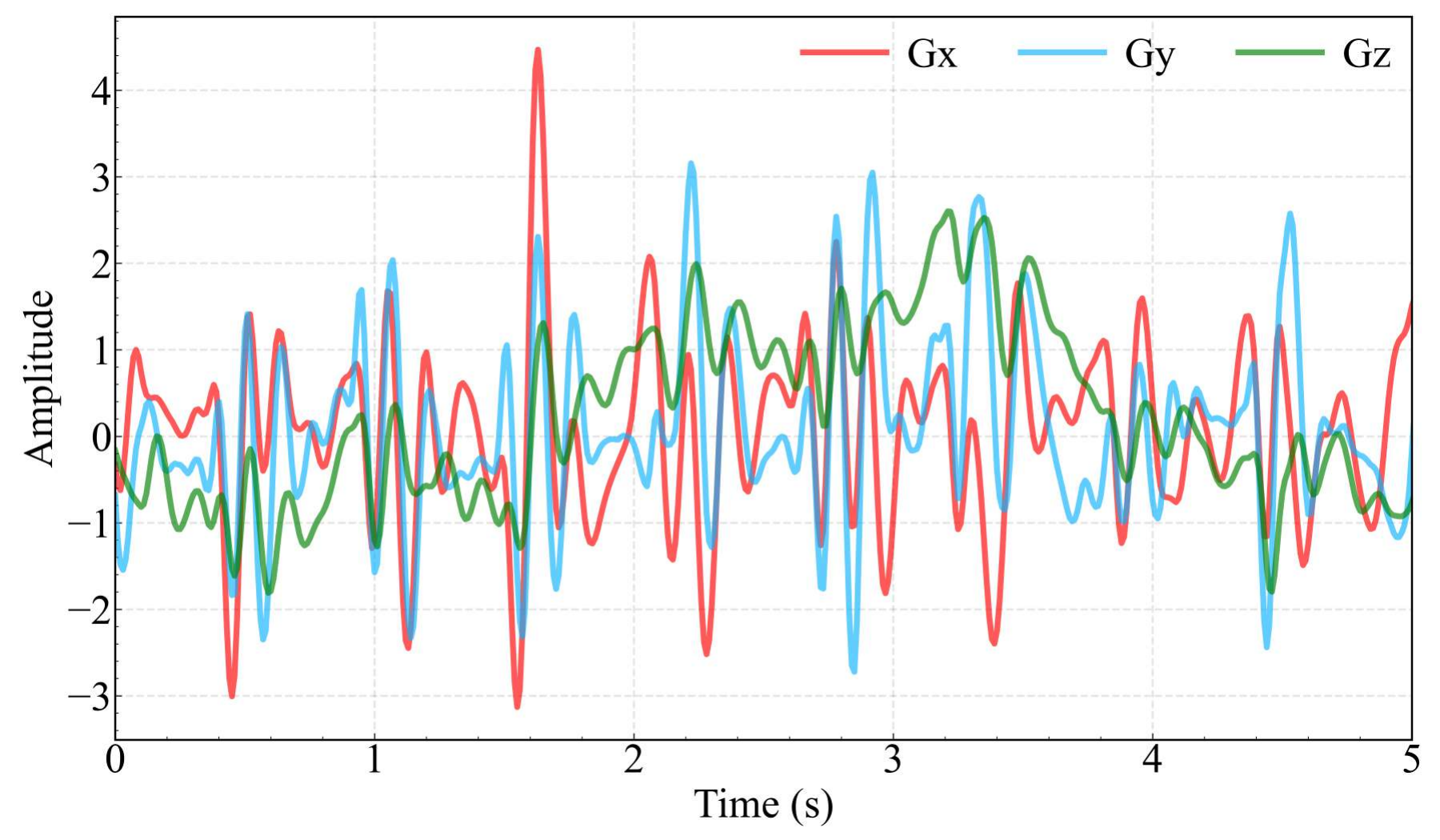}
    }
   \subfloat[Indoor turn around magnetometers]{
    \label{fig:indoor3}
    \includegraphics[width=0.3\textwidth]{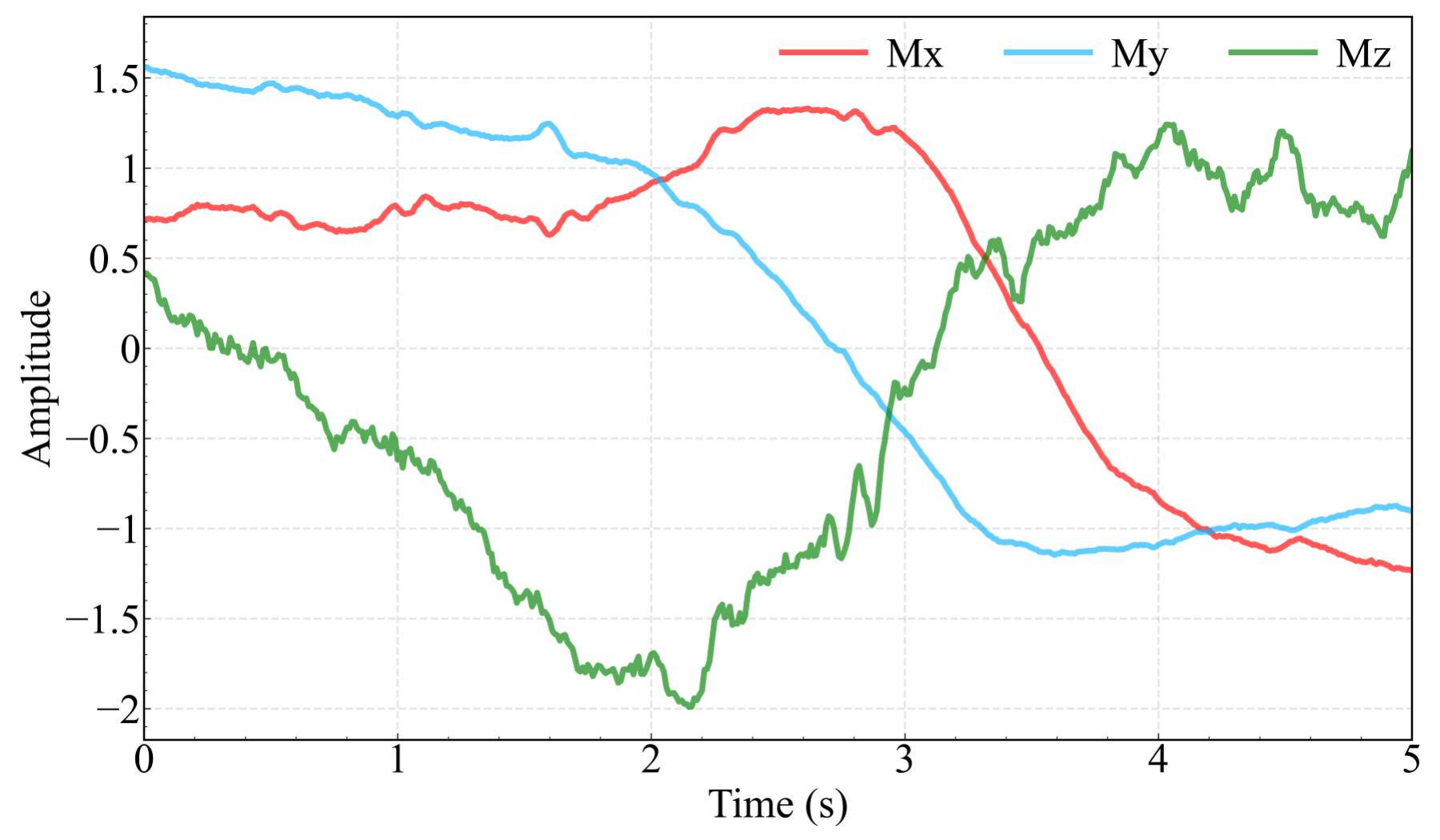}
    }\\
   \subfloat[Outdoor turn right accelerations]{
    \label{fig:outdoor1}
    \includegraphics[width=0.3\textwidth]{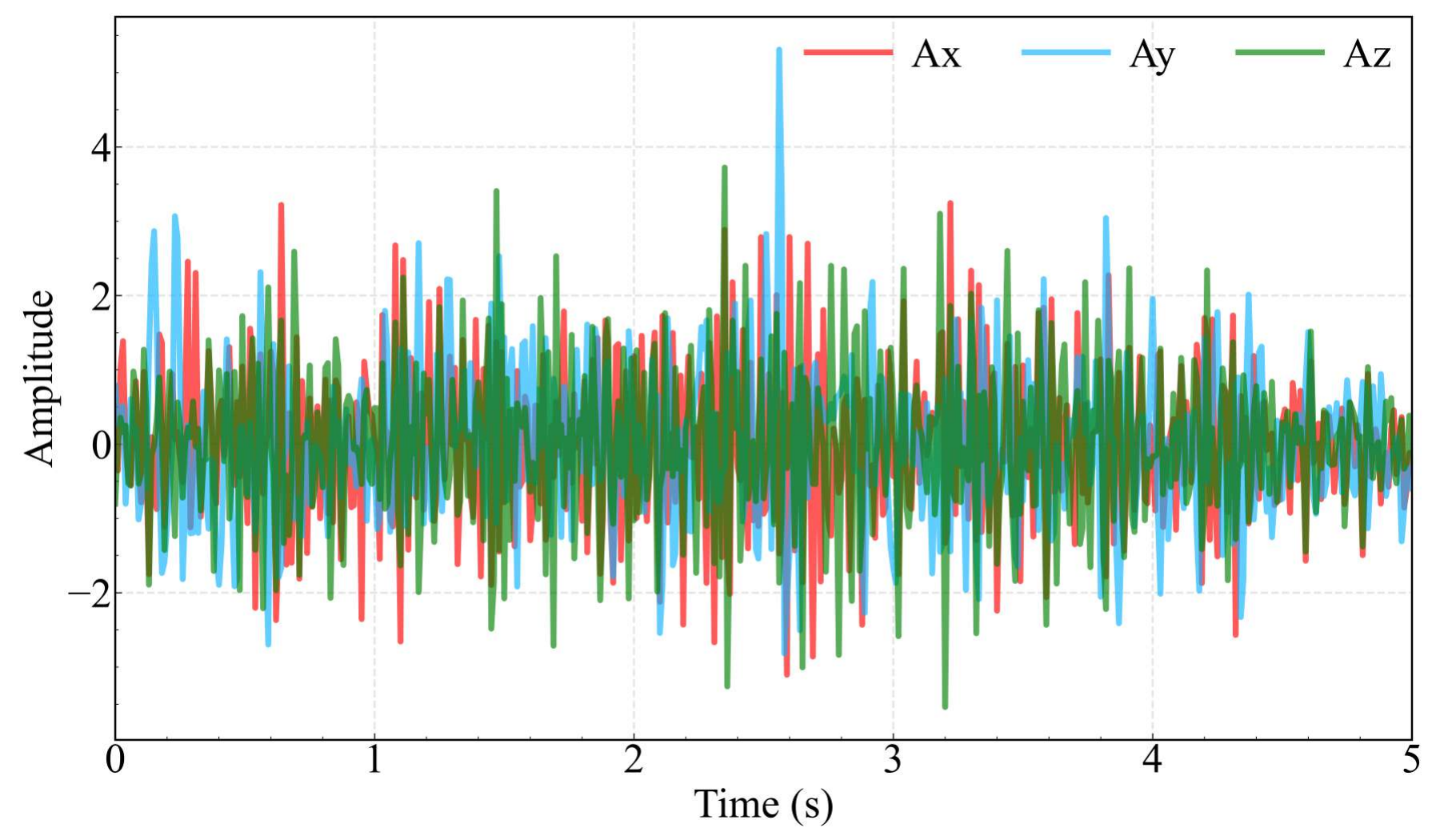}
    }
   \subfloat[Outdoor turn right gyroscopes]{
    \label{fig:outdoor2}
    \includegraphics[width=0.3\textwidth]{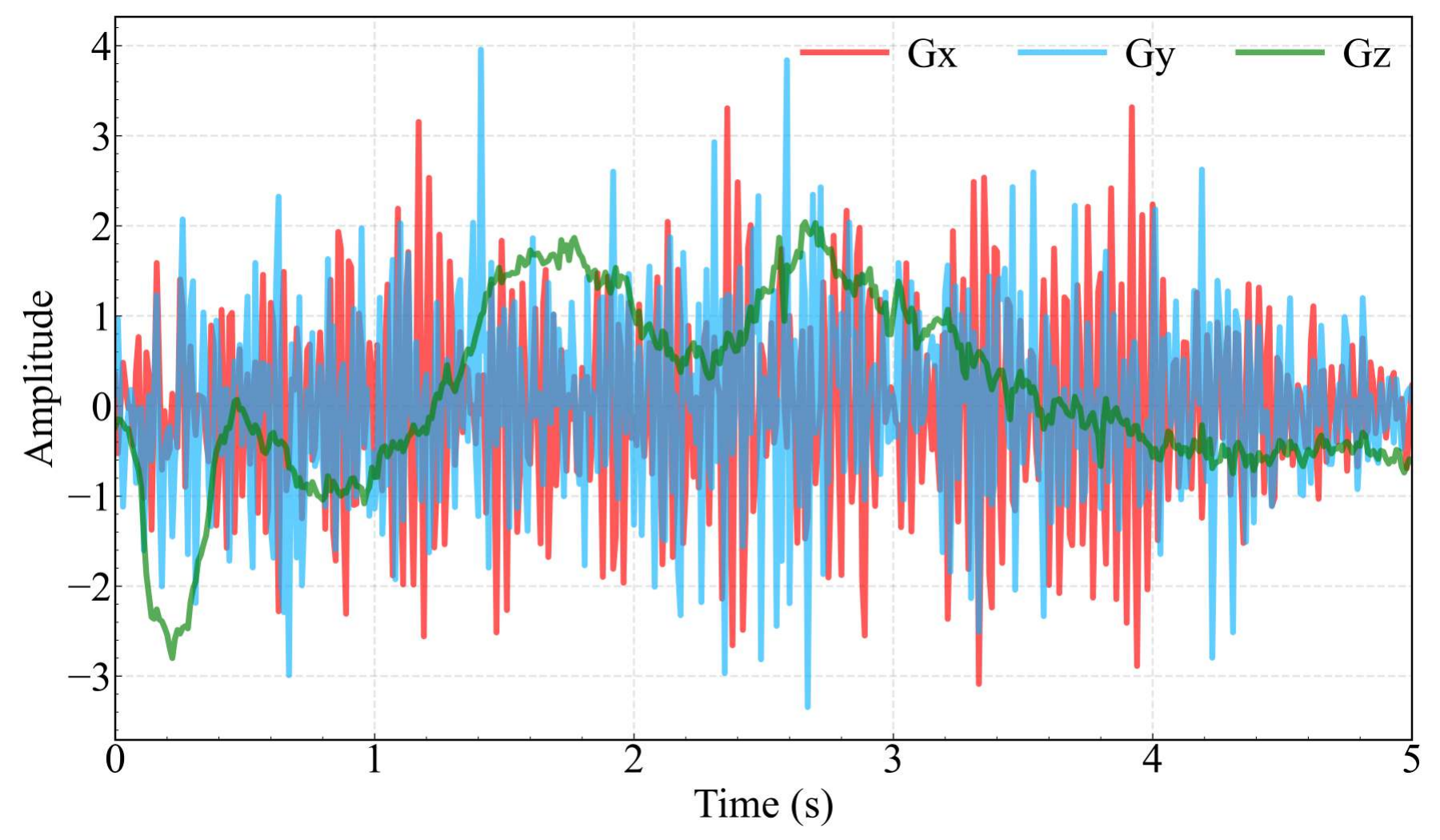}
    }
   \subfloat[Outdoor turn right magnetometers]{
    \label{fig:outdoor3}
    \includegraphics[width=0.3\textwidth]{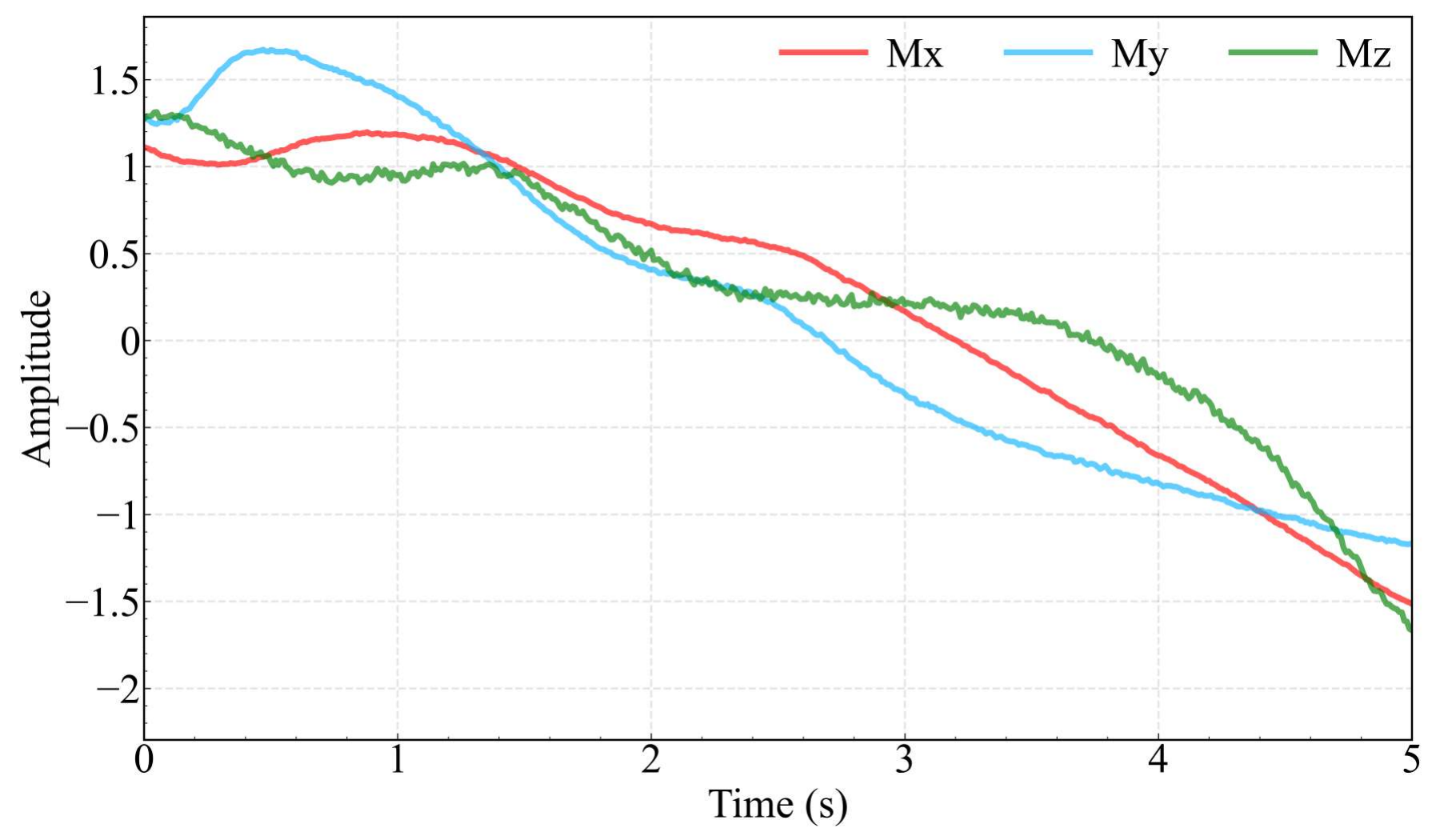}
    }
   \caption{IMU data visualization of two scenes including indoor turn right and outdoor turn right.}
   \label{fig:imu_data}
   \vspace{-0.5cm}
\end{figure*}
To examine the proficiency of LLMs in identifying robotic trajectories from raw Inertial Measurement Unit (IMU) data, we structured our investigation into two experimental stages, each tailored to different scenarios.
The preliminary stage focused on ascertaining the capability of LLMs to recognize trajectories in an indoor setting, characterized by uncomplicated navigational paths.
Moving to the second stage, the aim was to determine the extent to which these models could discern between trajectories within outdoor environments, which are not only complex due to the irregular ground surfaces but also notoriously difficult to distinguish, presenting a substantial challenge.

\subsubsection{Dataset Setup}
To support these experiments, we employed two distinct datasets, one captured indoors and the other outdoors, both collected by an autonomous robot equipped with a smartphone, as depicted in Fig.~\ref{fig:exp}. Additional information is detailed in the following.

\head{Indoor setting}
The indoor dataset encompasses a diverse array of robotic trajectories traversing inside a building, meticulously recorded via an IMU-equipped smartphone. For every trajectory, we harnessed the 9-axis IMU dataset, which is annotated with explicit labels: 'straight', 'turn right', 'turn left', and 'turn around'. An illustration of the sample raw data is showcased in Fig.~\ref{fig:indoor1} to Fig.~\ref{fig:indoor3}. This IMU consistently operates at a default sampling rate of 100Hz.
To benchmark the performance of LLMs against baseline models that require training data, we divided our dataset into training, validation, and seen test and unseen test sets at a ratio of 3:1:1:1. 
Furthermore, considering the necessity to feed raw IMU data as tokens into LLMs, we performed a downsampling of the IMU data to a frequency of 3Hz specifically for LLM inputs.

\head{Outdoor setting}
The outdoor dataset comprises various robot trajectories collected around a campus environment, utilizing 9-axis IMU data from smartphones mounted on top of the robots. We chose four trajectories as shown in Fig.\ref{fig:exp}. An illustration of the sample raw data is showcased in Fig.~\ref{fig:outdoor1} to Fig.~\ref{fig:outdoor3}.
Similar to the partitioning approach used for the indoor dataset, the outdoor dataset is also segmented into distinct sets for training, validation, and testing (both seen and unseen) at a ratio of 3:1:1:1. To accommodate the input requirements of LLMs, the IMU data is further downsampled to 3Hz before being tokenized for model processing.

\subsubsection{Baseline}
As benchmarks for comparison, we employ traditional machine learning models as well as deep learning models. These models require a training phase to develop their capacity for classification.

\head{Random Forest~\cite{biau2016random}}
Random forests (RF) is an ensemble learning technique used for classification tasks. It builds numerous decision trees during the training process and is known for its ability to effectively manage classes of varying sizes within classification problems. For our implementation, we utilize the Random Forest model from the scikit-learn library, applying the default parameters provided.

\head{Support Vector Machine~\cite{hearst1998support}}
The support vector machine (SVM) is a supervised learning model designed for both classification and regression tasks. It is known for its capacity to find the maximum margin in a dataset, often resulting in robust classification boundaries. Our experiments leverage the SVM implementation from scikit-learn, specifically using Gaussian kernels.

\head{CNN~\cite{yang2015deep}}
The Convolutional Neural Network is a deep learning architecture that excels in automatically extracting features from multichannel time series signals, such as those gathered by sensors on a robot. CNN is adept at learning intricate features from trajectory data and often outperform traditional approaches in classification accuracy.

\head{LSTM~\cite{xu2021limu}}
Long Short-Term Memory networks (LSTMs) are a specialized type of model designed explicitly for tasks requiring the understanding and memory of temporal sequences. Standing out from other foundational models, LSTMs handle input features in a sequential manner through time. This sequential processing capability allows LSTMs to excel in applications where the temporal dimension is critical, such as speech recognition~\cite{irie2016lstm}, channel forecasting~\cite{yang2022vehicle}, and gesture recognition~\cite{xu2024washring}.

\subsubsection{Prompt Structure}
LLMs have been recognized for their proficiency in few-shot learning, wherein they utilize prompts structured text or templates—to steer their output towards specific tasks~\cite{wei2022chain}. For instance, Nori et al. reported leveraging instructional cues to tap into the extensive medical expertise embedded within LLMs, setting new benchmarks across various datasets~\cite{nori2023can}. However, the limitation of this prompt-based approach is the requirement of crafting tailored question-and-answer frameworks, which can confine LLMs to narrowly-defined tasks~\cite{brown2020language}. Additionally, it has been observed that such tailor-made prompts can disrupt the natural inferencing process known as the Chain-of-Thought (CoT), potentially degrading performance~\cite{kojima2022large}.
To address these limitations, our strategy involves constructing prompts that are fundamentally simple, employing role-playing and incremental thought processes to naturally elicit the inherent capabilities of LLMs without depending on predefined answer frameworks.


\begin{figure}[t]
  \begin{center}
  \includegraphics[width=0.48\textwidth]{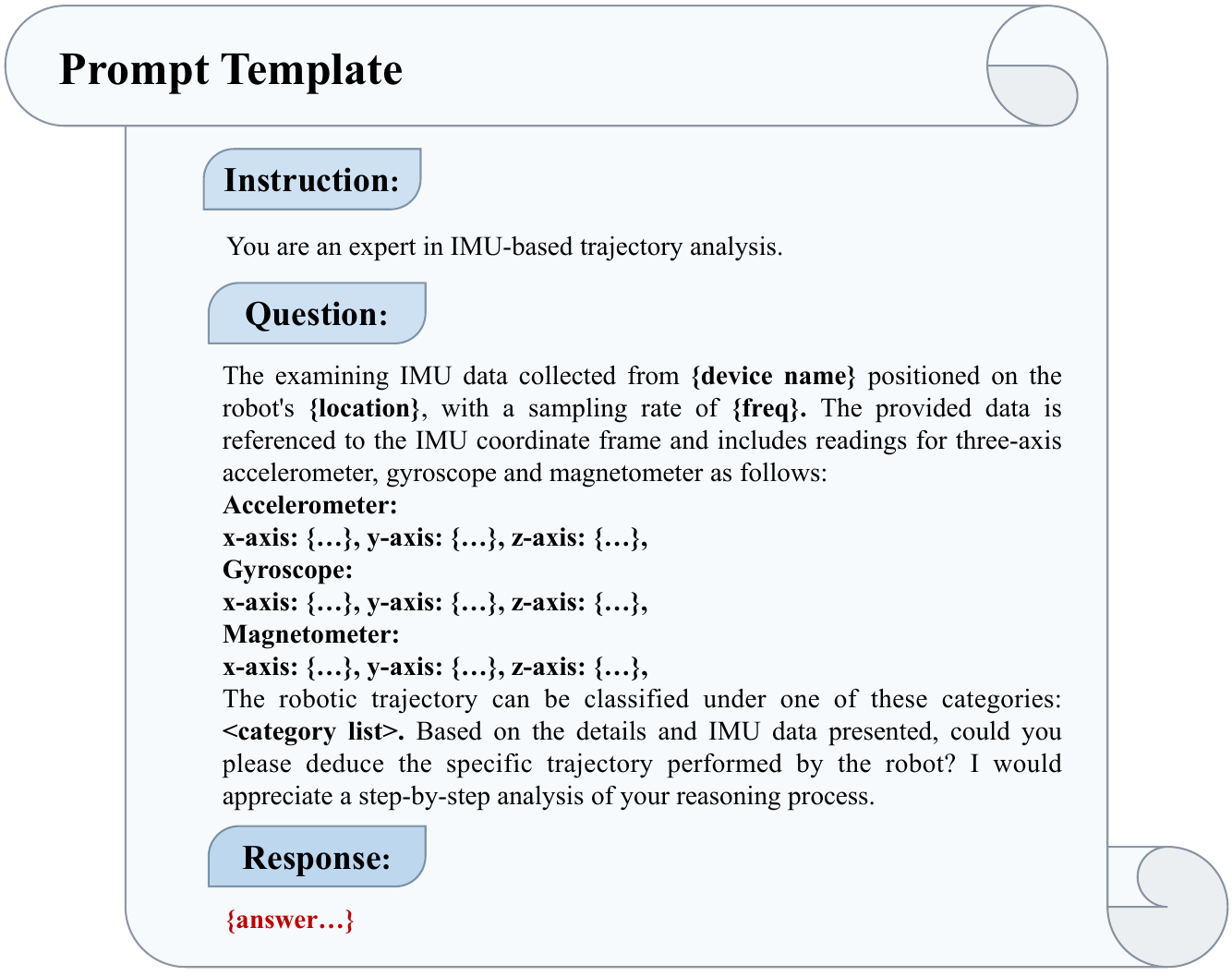}
  \caption{Chain-of-thought prompt design for LLMTrack.}\label{fig:prompt}
  \end{center}
  \vspace{-0.4cm}
\end{figure}

\begin{figure*}[h]
  \begin{center}
  \includegraphics[width=0.85\textwidth]{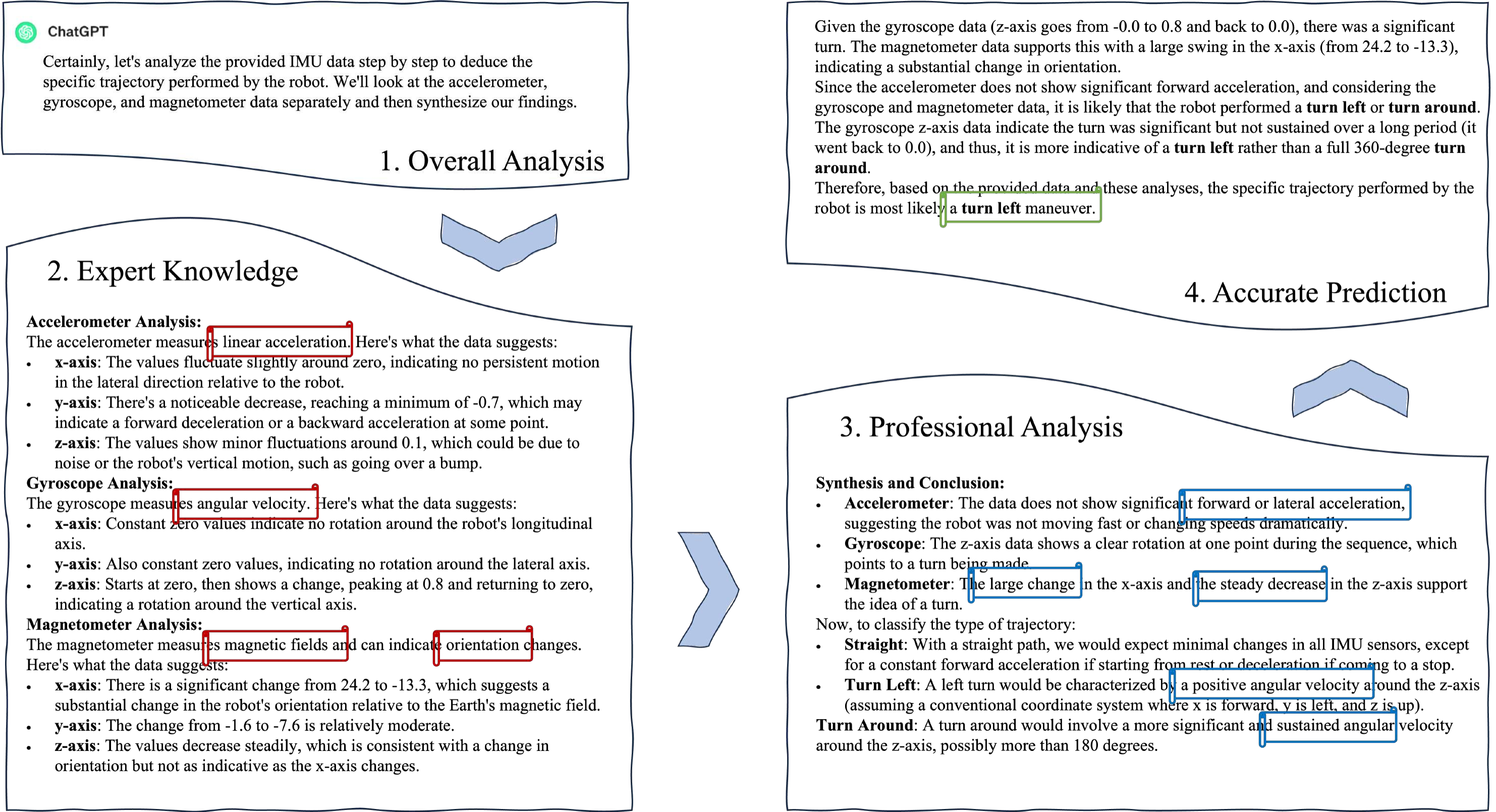}
  \caption{Detailed step-by-step inference generated by GPT4 with a turn-left example.}\label{fig:gpt4inference}
  \end{center}
\end{figure*}

\begin{table*}[t]
\centering
\caption{Overall performance. (\CIRCLE \thinspace for chosen, \Circle \thinspace for unchosen, $\rm{DO^*}$: direct output; $\rm{CoT^*}$: chain-of-thought.)}
\scriptsize
\setlength{\tabcolsep}{8pt} 
\begin{tabular}{l|cc|cc|ccc}
\toprule
\multicolumn{1}{c|}{\textbf{Models}} & \multicolumn{2}{c|}{\textbf{Scenarios}} & \multicolumn{2}{c|}{\textbf{Test subject}} & \multicolumn{3}{c}{\textbf{Accuracy}} \\
\cmidrule(lr){2-3}
\cmidrule(lr){4-5}
\cmidrule(lr){6-8}

& \textbf{Indoor} & \textbf{Outdoor} & \textbf{Seen} & \textbf{Unseen} & \textbf{Precision} & \textbf{Recall} & \textbf{F1-Score} \\

\midrule
\multirow{4}{*}{RF} & \CIRCLE & \Circle & \CIRCLE & \Circle & \timebar{87}{75.2}\% & \timebar{87}{68.2}\% & \timebar{87}{72.1}\% \\
                     & \CIRCLE & \Circle & \Circle & \CIRCLE & \timebar{87}{46.7}\% & \timebar{87}{42.3}\% & \timebar{87}{44.5}\% \\
                     & \Circle & \CIRCLE & \CIRCLE & \Circle & \timebar{87}{71.3}\% & \timebar{87}{66.2}\% & \timebar{87}{69.5}\% \\
                     & \Circle & \CIRCLE & \Circle & \CIRCLE & \timebar{87}{40.3}\% & \timebar{87}{44.9}\% & \timebar{87}{43.7}\% \\

\midrule
\multirow{4}{*}{SVM} & \CIRCLE & \Circle & \CIRCLE & \Circle & \timebar{87}{73.1}\% & \timebar{87}{71.5}\% & \timebar{87}{72.4}\% \\
                     & \CIRCLE & \Circle & \Circle & \CIRCLE & \timebar{87}{42.2}\% & \timebar{87}{45.1}\% & \timebar{87}{43.7}\% \\
                     & \Circle & \CIRCLE & \CIRCLE & \Circle & \timebar{87}{68.7}\% & \timebar{87}{72.3}\% & \timebar{87}{70.2}\% \\
                     & \Circle & \CIRCLE & \Circle & \CIRCLE & \timebar{87}{41.6}\% & \timebar{87}{46.2}\% & \timebar{87}{43.1}\% \\
\midrule
\multirow{4}{*}{CNN} & \CIRCLE & \Circle & \CIRCLE & \Circle & \timebar{87}{72.5}\% & \timebar{87}{76.4}\% & \timebar{87}{74.4}\% \\
                     & \CIRCLE & \Circle & \Circle & \CIRCLE & \timebar{87}{52.1}\% & \timebar{87}{40.8}\% & \timebar{87}{45.9}\% \\
                     & \Circle & \CIRCLE & \CIRCLE & \Circle & \timebar{87}{77.3}\% & \timebar{87}{68.6}\% & \timebar{87}{73.8}\% \\
                     & \Circle & \CIRCLE & \Circle & \CIRCLE & \timebar{87}{43.7}\% & \timebar{87}{59.2}\% & \timebar{87}{50.6}\% \\
\midrule
\multirow{4}{*}{LSTM} & \CIRCLE & \Circle & \CIRCLE & \Circle & \timebar{87}{78.2}\% & \timebar{87}{73.9}\% & \timebar{87}{76.1}\% \\
                     & \CIRCLE & \Circle & \Circle & \CIRCLE & \timebar{87}{57.8}\% & \timebar{87}{61.4}\% & \timebar{87}{59.5}\% \\
                     & \Circle & \CIRCLE & \CIRCLE & \Circle & \timebar{87}{75.1}\% & \timebar{87}{65.6}\% & \timebar{87}{70.3}\% \\
                     & \Circle & \CIRCLE & \Circle & \CIRCLE & \timebar{87}{52.7}\% & \timebar{87}{56.3}\% & \timebar{87}{54.4}\% \\

\midrule
\multirow{2}{*}{GPT4 - DO*} & \CIRCLE & \Circle & \Circle & \CIRCLE & \timebar{87}{55.7}\% & \timebar{87}{62.1}\% & \timebar{87}{59.1}\% \\
                     & \Circle & \CIRCLE & \Circle & \CIRCLE & \timebar{87}{57.2}\% & \timebar{87}{59.3}\% & \timebar{87}{58.2}\% \\

\midrule
\multirow{2}{*}{GPT4 - CoT*} & \CIRCLE & \Circle & \Circle & \CIRCLE & \timebar{87}{83.3}\% & \timebar{87}{80.1}\% & \timebar{87}{83.6}\% \\
                     & \Circle & \CIRCLE & \Circle & \CIRCLE & \timebar{87}{79.2}\% & \timebar{87}{74.5}\% & \timebar{87}{76.7}\% \\

\bottomrule
\end{tabular}
\label{tbl:type2}
\end{table*}

The construction of our prompt, as visualized in Fig.~\ref{fig:prompt}, is elegantly minimalist, comprising solely of an initial instruction followed by a probing question.
The directive is designed to activate the LLM's reservoir of specialized knowledge concerning IMU technology. Simultaneously, the question furnishes precise context about the data acquisition process, the downsampled sequence of raw IMU data, and the plausible categories of trajectories being investigated.
Our prompt culminates with an invitation for the LLM to "I would appreciate a step-by-step analysis of your reasoning process" a tactic intended to provoke a comprehensive CoT progression within the LLM. This method is championed by existing research~\cite{kim2024health,wei2022chain}, for its efficacy in refining the precision of LLM outputs.
As illustrated in Fig.~\ref{fig:gpt4inference}, by foregoing a restricted response format, we empower the Large Language Model (LLM) to freely generate a richer textual output. This approach leverages the model's inherent analytical capabilities to tap into and utilize the pertinent knowledge it has internalized.

\subsection{Evaluation}
We select GPT4~\cite{achiam2023gpt}, the most advanced and powerful LLM currently accessible, to conduct a comprehensive analysis with four other baseline models across two datasets.
Additionally, we compare the prompt mode of direct output (DO) without analysis to verify the effectiveness of applying CoT prompt for LLMs to improve prediction accuracy.

\begin{figure*}[h]
\centering
   \subfloat[GPT3.5]{%
    \label{fig:gpt35}
      \includegraphics[width=0.3\textwidth]{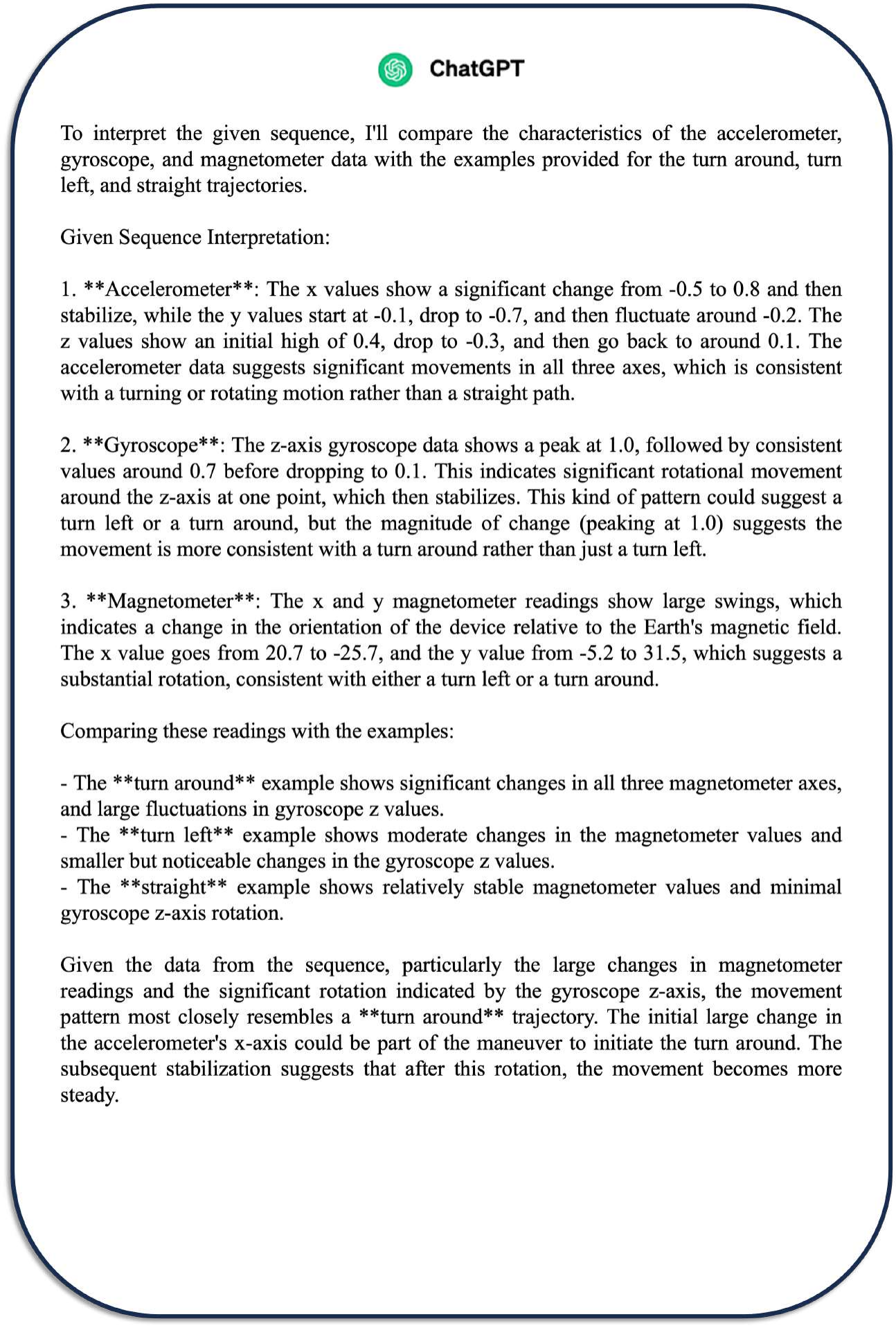}}%
   \subfloat[Claude]{%
    \label{fig:Claude}
      \includegraphics[width=0.3\textwidth]{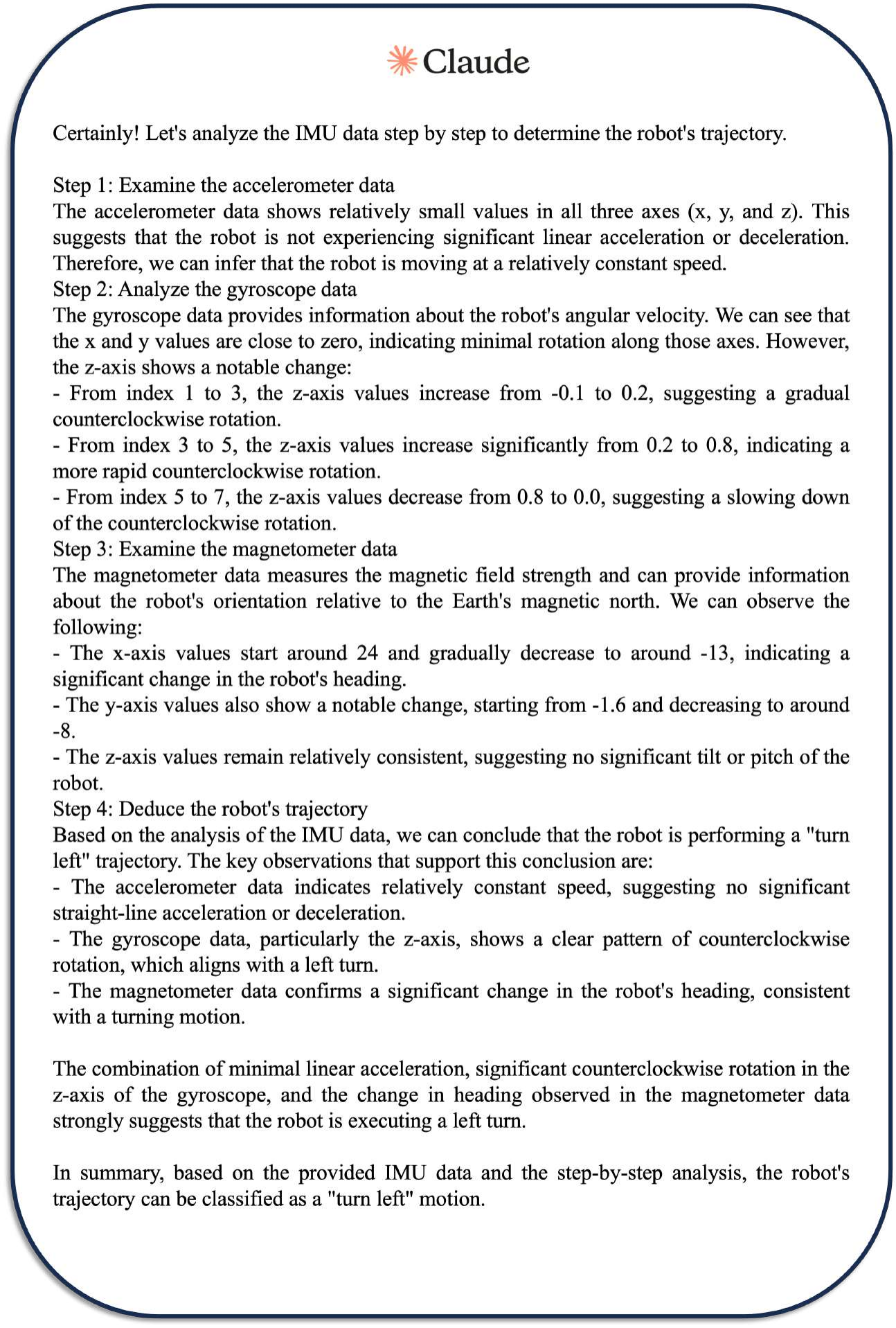}
	  }%
   \subfloat[Gemini]{%
    \label{fig:Gemini}
      \includegraphics[width=0.3\textwidth]{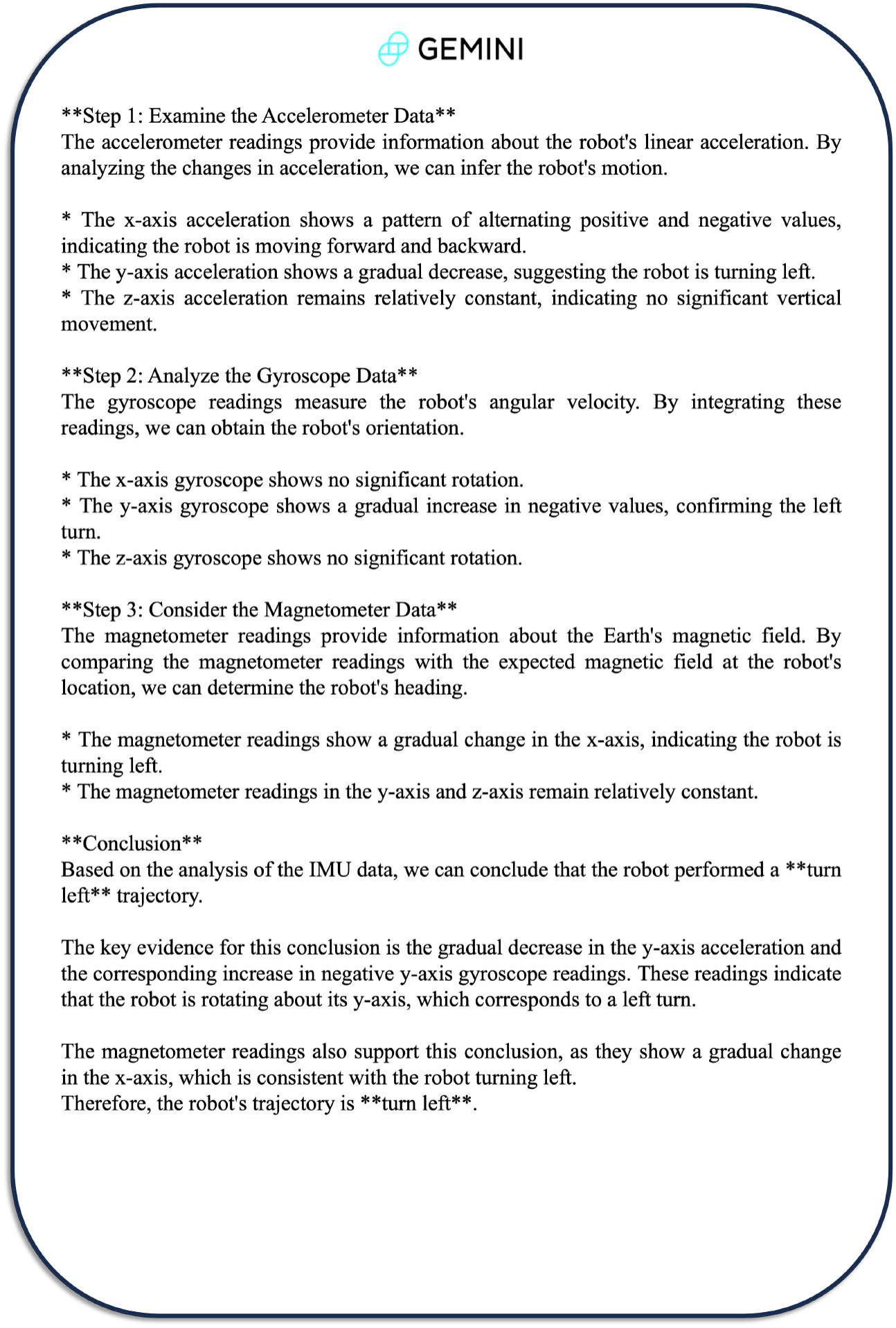}
	  }%
   \caption{A comparison of the inference results generated by other LLMs for the turn-left scenario.
}
   \label{fig:other-llms}
\end{figure*}

\head{Overall Performance}
The comprehensive test results are detailed in \tab\ref{tbl:type2}. GPT4, by harnessing its robust comprehension capabilities along with the CoT prompt strategy, has delivered outstanding performance across all baseline assessments on the unseen dataset. It boasts an impressive average F1-score of 80.1\%.
A deeper analysis reveals that in comparison to its predecessor, GPT4-DO, the GPT4 model enhanced with CoT prompts—referred to as GPT4-CoT—shows a remarkable improvement of 38\%. This significant enhancement underscores the value of contextualized prompting in achieving higher levels of model performance.
In contrast, traditional neural network models like LSTM and CNN, while considered top-tier baselines, only manage to achieve a relatively modest average F1-score of approximately 73\%. This suggests that while they remain useful, there is a substantial gap between their performance and the more advanced GPT4-CoT.
The comparison becomes even more stark when evaluating the performance of two other conventional machine learning methods, RF and SVM. These methods show markedly poorer performance, hovering around the 70\% mark. This performance disparity highlights the limitations of traditional machine learning techniques when faced with complex tasks that GPT4-CoT handles with relative ease.

\head{Performance under Different Scenarios}
We subsequently delved into an analysis of the model's performance across various environments. The data presented in Tab.~\ref{tbl:type2} confirms that, for nearly all models evaluated, indoor performance surpasses outdoor performance. More precisely, with the GPT4-DO model, indoor performance exceeds outdoor by 2\%. The discrepancy is even more pronounced with the GPT4-CoT model, where indoor results are 8\% higher than those outdoors. This discrepancy can largely be attributed to the outdoor environment's greater variability in floor surfaces, as opposed to the more uniform and flat indoor surfaces, which result in fewer fluctuations in the IMU sensor data.
In conclusion, the GPT4-CoT model demonstrates a robust capacity to achieve high performance in both indoor and outdoor scenarios, handling the varying conditions with commendable effectiveness.

\head{Detailed Inference Example}
To demonstrate the adeptness of GPT4 in generating expert-level insights and precise inferences, we provide a detailed representation of how GPT4 interprets the concept of a 'turn-left' maneuver in an indoor scenario, as illustrated in Fig.~\ref{fig:gpt4inference}.
The inference methodology is segmented into four distinct phases:
\begin{itemize}
    \item 
    Initially, GPT4 reviews and clarifies the provided information, ensuring a clear understanding of the problem at hand.
\item In the subsequent phase, GPT4 employs embedded "expert knowledge" to accurately interpret the raw IMU sensor data that corresponds to specific movements. For instance, both 'turn-left' and 'turn-right' maneuvers are associated with variations in the gyroscope readings, 'go-straight' exhibits minimal fluctuation, and 'turn-around' is characterized by more pronounced changes in gyroscope values.
\item The third phase involves GPT4 conducting a more granular analysis of the input raw data, which leads to the identification that the data is indicative of a particular pattern.
\item Finally, after a thorough analysis and integration of the previous expert insights, GPT4 arrives at a well-founded conclusion in the fourth stage, determining that the action is most likely a 'turn-left' trajectory.
\end{itemize}
This step-by-step process exemplifies GPT4's ability to not only process raw sensor data but also to apply specialized knowledge for accurate trajectory tracing.

\section{Discussion}
\head{Logical Reasoning Ability}
Previous studies have confirmed that LLMs are equipped with the potential to perform logical deductions. Moreover, our studies have revealed that their proficiency is not limited to textual data interpretation but also to making sense of data from sensory inputs.
Models like GPT4 have shown remarkable adaptability in translating raw sensory data into conceptual verbal constructs, identifying patterns as intermittent, static, or sudden. This capability is analogous to a powerful sieve that effectively tackles the issue of out-of-distribution data, a notable challenge for standard machine learning or deep learning models.
LLMs prioritize the creation of credible narratives that mirror human cognitive patterns, shifting away from a sole reliance on mimicking the exact features of raw inputs. Nonetheless, there are variances in the logical deductive strengths across different LLMs. The comparisons in \fig\ref{fig:other-llms} illustrate the variations in response to the same prompt when tested on other models such as GPT3.5, Anthropic Claude, and Google Gemini.
It is important to highlight that the presence of reasoning skills in a model does not directly imply higher intelligence. Notably, some smaller-scale LLMs like GPT3.5 and Google Gemini have shown to lag behind GPT4 in their reasoning accuracy, despite being prompted in the same way.


\section{Future Work and Conclusions}
\label{sec:conclusion}

The results of this research indicate that LLMs can serve as a primary framework for accurately and reliably mapping movement paths in a zero-shot context. This suggests that LLMs have the innate ability to interpret data from IoT sensors without relying on example-based guidance from experts. The study underscores the potential of LLMs to process unrefined sensor data, signaling a significant potential shift in the AIoT field. However, it is essential to conduct further research to determine the specific conditions and limits of LLM effectiveness. The establishment of a more rigorous and comprehensive set of evaluation methods and standards is necessary. Enhancing our understanding of the strengths and limitations of Language Models will enable us to utilize their full potential in analyzing complex real-world data, such as abstract Channel State Information in wireless devices.

\bibliographystyle{IEEEtran}

\bibliography{main}

\begin{thebibliography}{10}
\providecommand{\url}[1]{#1}
\csname url@samestyle\endcsname
\providecommand{\newblock}{\relax}
\providecommand{\bibinfo}[2]{#2}
\providecommand{\BIBentrySTDinterwordspacing}{\spaceskip=0pt\relax}
\providecommand{\BIBentryALTinterwordstretchfactor}{4}
\providecommand{\BIBentryALTinterwordspacing}{\spaceskip=\fontdimen2\font plus
\BIBentryALTinterwordstretchfactor\fontdimen3\font minus \fontdimen4\font\relax}
\providecommand{\BIBforeignlanguage}[2]{{%
\expandafter\ifx\csname l@#1\endcsname\relax
\typeout{** WARNING: IEEEtran.bst: No hyphenation pattern has been}%
\typeout{** loaded for the language `#1'. Using the pattern for}%
\typeout{** the default language instead.}%
\else
\language=\csname l@#1\endcsname
\fi
#2}}
\providecommand{\BIBdecl}{\relax}
\BIBdecl

\bibitem{li2024personal}
Y.~Li, H.~Wen, W.~Wang, X.~Li, Y.~Yuan, G.~Liu, J.~Liu, W.~Xu, X.~Wang, Y.~Sun \emph{et~al.}, ``Personal llm agents: Insights and survey about the capability, efficiency and security,'' \emph{arXiv preprint arXiv:2401.05459}, 2024.

\bibitem{zhang2024large}
X.~Zhang, R.~R. Chowdhury, R.~K. Gupta, and J.~Shang, ``Large language models for time series: A survey,'' 2024.

\bibitem{li2024survey}
Y.~Li, Z.~Li, P.~Wang, J.~Li, X.~Sun, H.~Cheng, and J.~X. Yu, ``A survey of graph meets large language model: Progress and future directions,'' 2024.

\bibitem{videoworldsimulators2024}
\BIBentryALTinterwordspacing
T.~Brooks, B.~Peebles, C.~Homes, W.~DePue, Y.~Guo, L.~Jing, D.~Schnurr, J.~Taylor, T.~Luhman, E.~Luhman, C.~Ng, R.~Wang, and A.~Ramesh, ``Video generation models as world simulators,'' 2024. [Online]. Available: \url{https://openai.com/research/video-generation-models-as-world-simulators}
\BIBentrySTDinterwordspacing

\bibitem{claude2023}
\BIBentryALTinterwordspacing
Claude. (2024, March). [Online]. Available: \url{https://www.anthropic.com/news/claude-3-family}
\BIBentrySTDinterwordspacing

\bibitem{lecun2022path}
Y.~LeCun, ``A path towards autonomous machine intelligence version 0.9. 2, 2022-06-27,'' \emph{Open Review}, vol.~62, 2022.

\bibitem{ji2024hargpt}
S.~Ji, X.~Zheng, and C.~Wu, ``Hargpt: Are llms zero-shot human activity recognizers?'' 2024.

\bibitem{biau2016random}
G.~Biau and E.~Scornet, ``A random forest guided tour,'' \emph{Test}, vol.~25, pp. 197--227, 2016.

\bibitem{hearst1998support}
M.~A. Hearst, S.~T. Dumais, E.~Osuna, J.~Platt, and B.~Scholkopf, ``Support vector machines,'' \emph{IEEE Intelligent Systems and their applications}, vol.~13, no.~4, pp. 18--28, 1998.

\bibitem{yang2015deep}
J.~Yang, M.~N. Nguyen, P.~P. San, X.~Li, and S.~Krishnaswamy, ``Deep convolutional neural networks on multichannel time series for human activity recognition.'' in \emph{Ijcai}, vol.~15.\hskip 1em plus 0.5em minus 0.4em\relax Buenos Aires, Argentina, 2015, pp. 3995--4001.

\bibitem{xu2021limu}
H.~Xu, P.~Zhou, R.~Tan, M.~Li, and G.~Shen, ``Limu-bert: Unleashing the potential of unlabeled data for imu sensing applications,'' in \emph{Proceedings of the 19th ACM Conference on Embedded Networked Sensor Systems}, 2021, pp. 220--233.

\bibitem{irie2016lstm}
K.~Irie, Z.~T{\"u}ske, T.~Alkhouli, R.~Schl{\"u}ter, H.~Ney \emph{et~al.}, ``Lstm, gru, highway and a bit of attention: An empirical overview for language modeling in speech recognition,'' in \emph{Interspeech}, 2016, pp. 3519--3523.

\bibitem{yang2022vehicle}
H.~Yang, H.~Liu, C.~Luo, Y.~Wu, W.~Li, A.~Y. Zomaya, L.~Song, and W.~Xu, ``Vehicle-key: A secret key establishment scheme for lora-enabled iov communications,'' in \emph{IEEE ICDCS}, 2022.

\bibitem{xu2024washring}
W.~Xu, H.~Yang, J.~Chen, C.~Luo, J.~Zhang, Y.~Zhao, and W.~J. Li, ``Washring: An energy-efficient and highly accurate handwashing monitoring system via smart ring,'' \emph{IEEE TMC}, 2024.

\bibitem{wei2022chain}
J.~Wei, X.~Wang, D.~Schuurmans, M.~Bosma, F.~Xia, E.~Chi, Q.~V. Le, D.~Zhou \emph{et~al.}, ``Chain-of-thought prompting elicits reasoning in large language models,'' \emph{Advances in Neural Information Processing Systems}, vol.~35, pp. 24\,824--24\,837, 2022.

\bibitem{nori2023can}
H.~Nori, Y.~T. Lee, S.~Zhang, D.~Carignan, R.~Edgar, N.~Fusi, N.~King, J.~Larson, Y.~Li, W.~Liu \emph{et~al.}, ``Can generalist foundation models outcompete special-purpose tuning? case study in medicine,'' \emph{arXiv preprint arXiv:2311.16452}, 2023.

\bibitem{brown2020language}
T.~Brown, B.~Mann, N.~Ryder, M.~Subbiah, J.~D. Kaplan, P.~Dhariwal, A.~Neelakantan, P.~Shyam, G.~Sastry, A.~Askell \emph{et~al.}, ``Language models are few-shot learners,'' \emph{Advances in neural information processing systems}, vol.~33, pp. 1877--1901, 2020.

\bibitem{kojima2022large}
T.~Kojima, S.~S. Gu, M.~Reid, Y.~Matsuo, and Y.~Iwasawa, ``Large language models are zero-shot reasoners,'' \emph{Advances in neural information processing systems}, vol.~35, pp. 22\,199--22\,213, 2022.

\bibitem{kim2024health}
Y.~Kim, X.~Xu, D.~McDuff, C.~Breazeal, and H.~W. Park, ``Health-llm: Large language models for health prediction via wearable sensor data,'' \emph{arXiv preprint arXiv:2401.06866}, 2024.

\bibitem{achiam2023gpt}
J.~Achiam, S.~Adler, S.~Agarwal, L.~Ahmad, I.~Akkaya, F.~L. Aleman, D.~Almeida, J.~Altenschmidt, S.~Altman, S.~Anadkat \emph{et~al.}, ``Gpt-4 technical report,'' \emph{arXiv preprint arXiv:2303.08774}, 2023.

\end{thebibliography}

\end{document}